# Leveraging Time Series Data in Similarity Based Healthcare Predictive Models: The Case of Early ICU Mortality Prediction

*Full Paper*

**Mohammad Amin Morid**
David Eccles School of Business
University of Utah
amin.morid@business.utah.edu

**Olivia R. Liu Sheng**
David Eccles School of Business
University of Utah
olivia.sheng@business.utah.edu

**Samir Abdelrahman**
Department of Biomedical Informatics
University of Utah
samir.abdelrahman@utah.edu

## Abstract

Patient time series classification faces challenges in high degrees of dimensionality and missingness. In light of patient similarity theory, this study explores effective temporal feature engineering and reduction, missing value imputation, and change point detection methods that can afford similarity-based classification models with desirable accuracy enhancement. We select a piecewise aggregation approximation method to extract fine-grain temporal features and propose a minimalist method to impute missing values in temporal features. For dimensionality reduction, we adopt a gradient descent search method for feature weight assignment. We propose new patient status and directional change definitions based on medical knowledge or clinical guidelines about the value ranges for different patient status levels, and develop a method to detect change points indicating positive or negative patient status changes. We evaluate the effectiveness of the proposed methods in the context of early Intensive Care Unit mortality prediction. The evaluation results show that the k-Nearest Neighbor algorithm that incorporates methods we select and propose significantly outperform the relevant benchmarks for early ICU mortality prediction. This study makes contributions to time series classification and early ICU mortality prediction via identifying and enhancing temporal feature engineering and reduction methods for similarity-based time series classification.

## Introduction

Patient time series data are collected over time at varying time intervals to update patient status and to support medical decisions, leading to a wide variety of patient time series data – e.g., vital signs, lab results, diagnoses, prescriptions and billings in Electronic Health Records (EHRs) and other healthcare information systems. Past studies have extracted and leveraged temporal patterns (e.g., temporal statistics, trends, transitions and similarity) from time series data for patient event, (e.g., readmission or mortality), risk (Johnson et al. 2012), cost prediction (Bertsimas et al. 2008), or performance prediction (Cho et al. 2008). Some of the past studies have reduced such problems to one of classifying one or multiple time series of the same entity into different outcome/decision classes, which is termed the time series classification (TSC) problem (Lee et al. 2012). Time series classification can be tackled by engineering temporal features that provide meaningful representations of time series and predictive power at reduced dimensionality for use with general-purposed classification methods (Hippisley-Cox et al. 2009). The synergy between a certain temporal abstraction approach and classification algorithm varies amongst the vast choice space, and hence must be properly considered. To diagnose new patients, physicians are often influenced by previous similar cases with relevant clinical evidences, resulting in patient similarity theory in the clinical decision support literature (Rouzbahman and Chignell 2014). In light of this theory, this study focuses on enhancing similarity-based classification methods for patient time series classification.

The challenges of time series data include high dimensionality and high degree of missingness. Temporal abstraction of time series, missing value imputation and feature reduction approaches provide options to address these challenges. In addition, changes or transitions in various contexts including time series classification have provided essential predictive power (Lin et al. 2014). To explore effective temporal abstraction, missing value imputation, feature weight assignment and change point detection methods that can afford the $k$-Nearest-Neighbor ($k$NN) algorithm with desirable synergy, this study asks these research questions:

*"What are the effective temporal abstraction, missing value imputation and feature reduction methods for similarity-based time series classification?"*

*"What is an effective approach to define and detect patient status change points for similarity-based time series classification?"*

# Proposed Framework for Similarity-based Patient Time Series Classification

In this section, we introduce the methods this study has selected or proposed for patient time series classification and some background and justifications for these methods.

## Similarity based classification

Similarity-based classifiers estimate the class label of a test or new sample based on the similarities between the test sample and a set of labeled training samples, and the pairwise similarities between the training samples (Chen et al. 2009). The most popular family of algorithms has grown out the $k$-Nearest-Neighbor ($k$-NN) algorithm where $k$ is the number of training samples with maximal similarities to a test sample. Many advancements of the original $k$-NN including ENN (two-way similarity) (Tang and He 2015), CNN (condensed nearest neighbors) (Hart 1968) and kernel-based approaches (e.g., SVM-KNN) have been proposed (Chen et al. 2009) for similarity-based classification as well. To examine the usefulness of temporal features and reduction, and their synergy with similarity-based classification approaches, we made a conscientious decision to use the original $k$-NN in this study. The potentials of advancing the methods for patient time-series classification based on other similarity-based classification remain as future research directions.

Past research on similarity measure has led to a wide variety of time series distance functions such as Dynamic Time Wrapping (DTW) (Berndt and Clifford 1994), Edit Distance with Real Penalty (ERP) (Chen and Ng 2004) and Longest Common Subsequence (LCSS) (Vlachos et al. 2002) to measure dissimilarities based on different considerations. Our empirical exploration of different distance functions shows that the simplicity of the Manhattan distance function offers desirable flexibility to leverage the joint benefits of distance function, PAA grain size and missing value imputation methods. We hence select the Manhattan distance function in $k$-NN for patient time series classification.

## Temporal abstraction

Temporal abstraction decides on how to transform time series data into the input features of a classification model. The challenges of time series data include high dimensionality and high amount of missing data amongst others. Extant temporal abstraction methods vary in their considerations to address these challenges and the predictive power of the resulting temporal features (Fu 2011). One commonly used simple temporal abstraction approach, called piecewise aggregation approximation (PAA), segments a time series into a sequence of fixed-sized non-overlapping consecutive windows (or intervals) (Lin et al. 2003). Each window is represented by the average of all data values time-stamped within the window. We regard the size of a PAA window as the grain size and divide the temporal features produced by PAA into fine-grain features versus coarse grain features. Table 1 compares the dimensionality, degrees of missingness and information loss of fine-grain versus coarse-grain features.

|  | **Fine-Grain** | **Coarse-Grain** |
|---|---|---|
| **Window size** | Small | Large |
| **Information loss** | Low | High |
| **Dimensionality** | High | Low |
| **Missingness** | High | Low |

**Table 1. Comparison of fine-grain versus coarse-grain PAA temporal abstraction**

The low information loss of fine-grain PAA could afford $k$-NN improved accuracy over coarse-grain PAA features. Exploring effective grain-size, dimensionality reduction and missing value imputation methods are necessary to realize additional predictive power of fine-grain PAA temporal features.

**PAA grain decision**

The optimal PAA grain decision is analytically intractable due to the apparent complexity of considering the interrelated factors including missing value imputation, feature reduction and distance function. Empirical comparison should be employed to decide on the grain size. To reduce the number of empirical experiments, we only compare the classification accuracy resulting from different grain sizes while holding other methods fixed at the selected or proposed settings for time series classification.

**Missing value imputation (MVI)**

Missing value handling methods such as propensity score imputation, predictive model–based imputation and hot-deck imputation can be found from past literature (Penny and Chesney 2006). Some of the time series distance functions such DTW also incorporates missing value imputation (Berndt and Clifford 1994). In particular, DTW considers prior and posterior values of a missing value at a time point when deciding on the similarity between two time series. Motivated by DTW, we propose an adjacency-based imputation method which replaces a missing value by its posterior value if its prior value is not available or by its prior value if its posterior value is not available. If both prior and posterior values are available, their average becomes the imputed value. The proposed imputation can be performed independent of the distance function of choice.

**Feature weighting (FW)**

Feature reduction for classification can utilize a variety of approaches such as information gain, Gini index and Chi-square metrics to calculate feature rankings or weights for feature selection or reduction (Singh et al. 2010). Because of the well-tested ability to improve accuracy, we adopt the Gradient Descent (GS) method (Modha and Spangler 2003; Wettschereck and Aha 1995) to assign weights to time series features.

*Change point detection (CPD)*

Many change point detection methods focus on detecting changes in the mean, variance or trend in a time series that follows a distribution – e.g., Gaussian, normal or regression (Hawkins and Zamba 2012). Such methods are not appropriate for detecting change points in patient time series due to the underlying data distribution assumptions. In addition, a change in the mean or variance of numeric patient time series, for example, of blood pressure may not be a change in patient status if both the values before and after a change point represent the same patient status – e.g. normal. Therefore, this paper uses a change point detection method based on clinical domain knowledge.

**Directional change point detection method**

Past research has emphasized the importance of change point in clinical guidelines and decision supports (Assareh et al. 2011; La Rosa et al. 2008; Sawaya et al. 2011). Few or many change points of a patient's time series can easily differentiate a patient's condition and outcomes. In this paper we propose to define and detect change points based on changes in the health status according to categories of values in the time series rather than measures like mean or variance. For instance, systolic blood pressure less than 120 is considered normal, while between 120 and 139 is considered Prehypertension (Le et al. 2013). A directional change point is defined as a change in a patients' status category.

We denote a directional change point of patient $i$ for time series (TS) $j$ as $DCP(i, j)$. The input is the $status(i, j, k)$ that indicates the health status level of patient $i$ during time window $k$ based on TS $j$. The status output is an integer, where the lowest value of status for a feature represents the worst category of health status, while the highest value of status for this feature represents the best category of health status. Status change of patient i at time window k in terms of TS j is computed as follow:

$StatusChange(i, j, k)$ = *Positive, if status(i, j, k) > status (i, j, m), m<k*
= *Negative, if status(i, j, k) < status (i, j, m), m<k*
= *Stable, if status(i, j, k) = status (i, j, m), m<k*

where m is the time window of the most recent available status of the patient before window $k$. We propose three change point features for patient $i$ based on TS $j$ - the number of directional change points or $Num\_of\_DCP(i, j)$, and the first and last status changes, or LastStatusChange(i,j), and FirstStatusChange(i,j). The following determines the value of directional change point of patient $i$ status at time windows $k$ in terms of TS $j$:

$DCP(i, j, k)$ = *1, If status(i, j, k) is opposite to status (i, j, r), r<k*
= *0, if status(i, j, k) = is not opposite to status (i, j, r), r<k*

where r is the most recent available status change of the patient before window $k$ that is either positive or negative (i.e., Stable is not counted). DCP(i,j,1) for the first window is set to zero. Assume W is the number of time windows (e.g., 24), the number of directional change points is derived as:

$$Num\_of\_DCP(i, j) = \sum_{k=1}^{W} DCP(i, j, k)$$

The most recent and most related study to this method is the change point detection method proposed by Lin et al. (2014)(Lin et al. 2014) for time-to-event prediction of chronic conditions using EHR data. To detect change in patients' status, the numerical time series values are replaced by three nominal states (i.e., high, medium, low) based on numerical trend and two nominal trends (i.e., decrease, increase, stable) based on numerical value changes of each predictor. Their change point detection will be used as a benchmark for evaluating our proposed domain based directional change point detection method.

# $k$NN-TSC-FIWC

We refer to the proposed patient time series classification method that combines the $k$NN algorithm with fine-grain temporal features (F), missing-value imputation (I), feature weight assignment (W) and change point detection (C) methods we select or propose to enhance similarity-based time series classification as $k$NN-TSC-FIWC. Figure 1 summarizes the flow of the training and testing phases of $k$NN-TSC-FIWC.

Another benchmark of $k$NN-TSC-FIWC is Lee et al. (2012)(Lee et al. 2012) which proposes a similarity based time series classification algorithm - KNN-TSC to predict customer churn after 30 days. KNN-TSC divides each feature's time series data into 15 equal size intervals and adopts the Discrete Fourier transform (DFT) technique for time series similarity calculation. It doesn't assign feature weights. KNN-TSC utilizes stratified average voting to estimate the churn decision of a test sample. In empirical evaluation, we will compare the accuracy of $k$NN-TSC-FIWC to that of KNN-TSC.

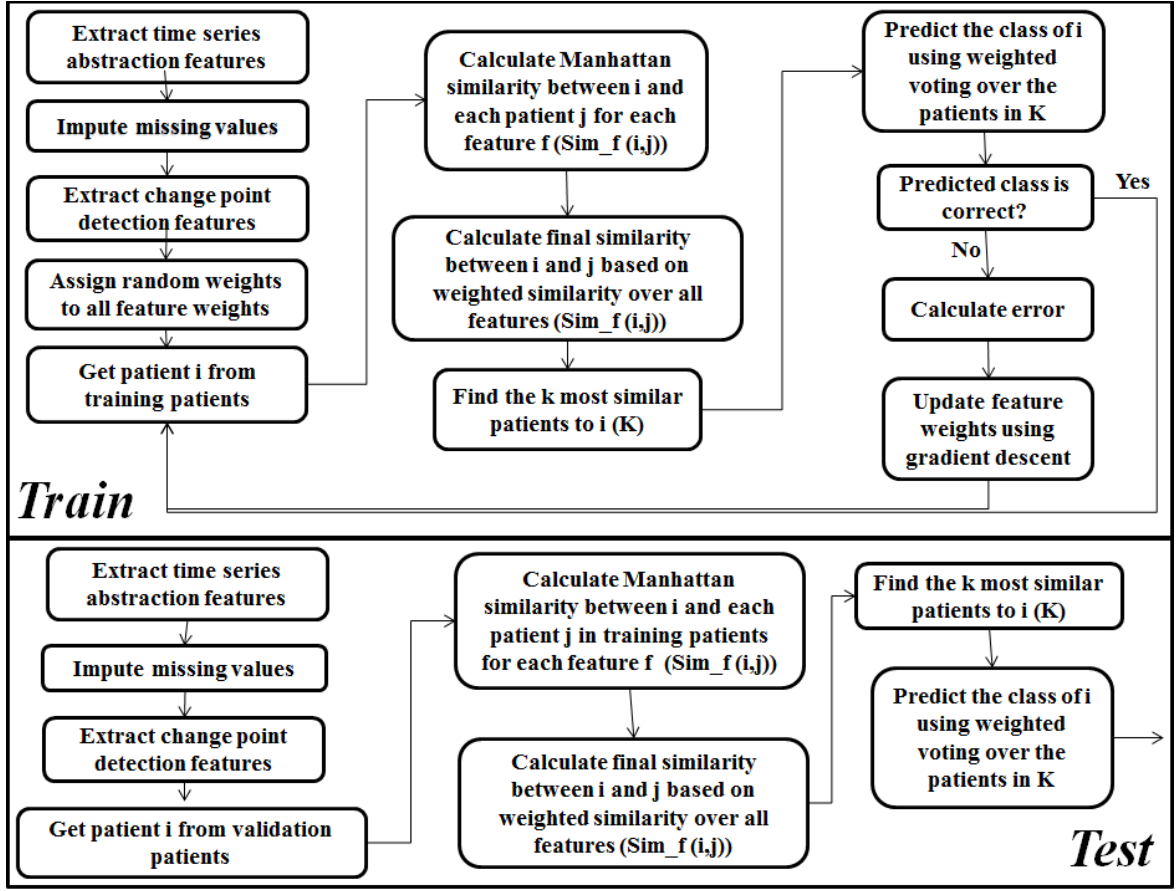

**Figure 1: An overview of kNN-TSC-FIWC**

## Empirical Evaluation

To evaluate the effectiveness of *k*NN-TSC-FIWC and its enabling methods, we compare them to benchmarks representing combinations of different temporal features, classification algorithms, similarity functions and dimension reduction methods in the context of early Intensive Care Unit (ICU) mortality prediction. Accurate ICU mortality prediction impacts medical therapy, triaging, end-of-life care, and many other aspects of ICU care (Gartman et al. 2009). To enhance the performance of ICU mortality prediction more sophisticated machine learning methods have been utilized recently. The PhysioNet/Computing in Cardiology 2012 Challenge aimed to provide a benchmark environment for early ICU mortality prediction (Silva et al. 2012). To the best of our knowledge, its winner (CCW hereafter) is the best early ICU mortality prediction benchmark using patients' first 48 hours of ICU time series data. CCW utilizes a new Bayesian ensemble scheme comprising of 500 weak decision tree learners which randomly assigns an intercept and gradient to a randomly selected single feature (Johnson et al. 2012).

### *Data and evaluation procedure*

To compare our results with CCW, we use the same experimental setup in the competition where patients were filtered to 22,561 patients who are younger than 16 years old and remained in the ICU for at least 48 hours. The data input consists of time series data of 36 variables (e.g., Glasgow Coma Score GCS) extracted from patients' ICU stay, plus four static features (i.e., age, gender, height, and initial weight). The target variable is a binary feature showing whether or not the patient eventually dies in the hospital before discharge. While almost half of the ICU patients have died eventually, most of the deaths happened out of hospital. The problem we analyze in this study is the prediction of in hospital mortality, which has an imbalanced distribution of 18% positive against 82% negative as shown in Table 2. It is interesting to observe that the difference in average ICU stays is very minor while the difference in outcomes is life versus death. Early mortality prediction may be able to help decision makers find ways to improve an ICU patient's survival rate. For finding and tuning the parameters including finding the best k, 50% of the data was used, while the rest remained unseen for validation. In all experiments, 20-fold cross validation was used to evaluate the performance of each method based on the validation dataset. Classification performance was measured according to the average precision, recall, and F-measure across 20 folds.

|  | **Number of Patients** | **Average Hospital Stay** | **Average ICU Stay** |
|---|---|---|---|
| **Alive** | 11977 (53%) | 9.68 | 5.21 |
| **Died** | 10584 (47%) | 13.87 | 5.53 |
| **Died out of Hospital (no)** | 6516 (29%) | 13.39 | 4.86 |
| **Died in Hospital (yes)** | 4068 (18%) | 13.5 | 6.6 |
| **Died in ICU (yes)** | 3240 (14%) | 10.53 | 7.49 |

**Table 2: Data distribution over the target variable (mortality)**

### Results

Table 3 shows that using two hours' time windows for the proposed method outperforms the same method with one, four and eight hours' time windows. The performance of the smallest grain size suffers from high missingness and the resulting noises in two patients' common PAA temporal values, while high loss of information details hurts the performance of large grain sizes. The best prediction performance is reached when the effect of missing values and information loss is balanced at window size of 2. Hence, the grain size chosen for the rest of the evaluations is 2 hours.

| | **Without change point detection ($k$NN-TSC-FIW)** | | | **With change point detection ($k$NN-TSC-FIWC)** | | |
|---|---|---|---|---|---|---|
| **Window size** | **Accuracy** | **F-Measure (yes)** | **F-Measure (no)** | **Accuracy** | **F-Measure (yes)** | **F-Measure (no)** |
| **1** | 0.72 | 0.56 | 0.81 | 0.80 | 0.69 | 0.91 |
| **2** | 0.78 | 0.66 | 0.89 | 0.82 | 0.77 | 0.93 |
| **4** | 0.75 | 0.63 | 0.76 | 0.80 | 0.69 | 0.91 |
| **8** | 0.74 | 0.61 | 0.83 | 0.75 | 0.63 | 0.86 |
| **12** | 0.66 | 0.48 | 0.75 | 0.64 | 0.45 | 0.74 |
| **24** | 0.55 | 0.30 | 0.7 | 0.55 | 0.30 | 0.70 |
| **48** | 0.44 | 0.19 | 0.62 | 0.43 | 0.16 | 0.60 |

**Table 3. Window size (abstraction size effect) effect on performance**

Table 4 compares the performance of $k$NN-TSC-FIW where a few benchmarking distance functions replace the Manhattan distance function. Although the performance results are close, they do validate the performance benefit the simple Manhattan distance function offers.

| | Without Change Point (*k*NN-TSC-FIW) | | | With Change Point (*k*NN-TSC-FIWC) | | |
|---|---|---|---|---|---|---|
| | Accuracy | F-Measure (yes) | F-Measure (no) | Accuracy | F-Measure (yes) | F-Measure (no) |
| **Manhattan** | 0.78 | 0.66 | 0.89 | 0.82 | 0.77 | 0.93 |
| **Euclidean** | 0.75 | 0.62 | 0.84 | 0.81 | 0.73 | 0.92 |
| **DTW** | 0.71 | 0.59 | 0.81 | 0.8 | 0.7 | 0.91 |
| **EDR** | 0.69 | 0.53 | 0.77 | 0.79 | 0.68 | 0.9 |
| **ERP** | 0.68 | 0.52 | 0.77 | 0.75 | 0.63 | 0.86 |
| **DFT** | 0.64 | 0.46 | 0.74 | 0.7 | 0.58 | 0.81 |
| **TSC** | 0.64 | 0.46 | 0.74 | 0.7 | 0.58 | 0.81 |
| **LCSS** | 0.57 | 0.33 | 0.71 | 0.66 | 0.48 | 0.75 |

**Table 4. Comparison of different time-series distance function**

Table 5 compares the performance of kNN-FIW against some of the well-established data mining methods, including support vector machine (SVM), the original kNN without feature weight assignment, neural network (NN) and logistic regression (LR). The input features for these algorithms are derived based on the fine-grain PAA of 2-hr window size. The comparison validates the performance advantage of similarity-based classification over its non-similarity counter-parts for early ICU mortality prediction.

Table 6 compares the performance of *k*NN combined with the selected or proposed methods starting with fine-grain time series abstraction (*k*NN-TSC-F), missing value imputation (*k*NN-TSC-FI), feature weighting (*k*NN-TSC-FIW) and change point detection (*k*NN-TSC-FIWC). The features based on the proposed fine-grain temporal abstraction, missing value imputation and feature weighting help the *k*NN-TSC-FIW model outperform the CCW benchmark by increasing the F-measure of the "yes" class by 11%. The proposed change point features also further double this performance improvement. The significance and benefits of these performance improvements in early ICU mortality by the proposed classification features cannot be underestimated.

Table 7 shows the significant effect of the proposed feature weighting technique on the proposed method (without considering change point features) against well-established feature weighting techniques, including Gini index, Chi-square and information gain, as well as the method proposed by Lee et al. (Lee et al. 2012).

| | Without Change Point | | | With Change Point | | |
|---|---|---|---|---|---|---|
| | Accuracy | F-Measure (yes) | F-Measure (no) | Accuracy | F-Measure (yes) | F-Measure (no) |
| ***k*NN-TSC-FIW (left) *k*NN-TSC-FIWC (right)** | 0.78 | 0.66 | 0.89 | 0.82 | 0.77 | 0.93 |
| **CCW** | 0.7 | 0.55 | 0.81 | 0.75 | 0.62 | 0.84 |
| **CCW on fine-grain features** | 0.66 | 0.48 | 0.75 | 0.71 | 0.59 | 0.81 |
| **SVM** | 0.65 | 0.47 | 0.85 | 0.7 | 0.58 | 0.81 |
| **LR** | 0.61 | 0.42 | 0.74 | 0.72 | 0.56 | 0.81 |
| **kNN** | 0.68 | 0.5 | 0.77 | 0.68 | 0.52 | 0.77 |
| **NN** | 0.64 | 0.46 | 0.74 | 0.68 | 0.52 | 0.77 |

**Table 5. Similarity based method against son-similarity based methods using fine-grain abstraction**

| Method | Grain | MVI | FW | CPD | Accuracy | F-Measure (yes) | F-Measure (no) |
|--------|-------|-----|-----|-----|----------|-----------------|----------------|
| **CCW** | Coarse | No | No | No | 0.70 | 0.55 | 0.81 |
| **kNN-TSC-F** | Fine | No | No | No | 0.55 | 0.29 | 0.71 |
| **kNN-TSC-FI** | Fine | Yes | No | No | 0.64 | 0.40 | 0.83 |
| **kNN-TSC-FIW** | Fine | Yes | Yes | No | 0.78 | 0.66 | 0.89 |
| **kNN-TSC-FIWC** | Fine | Yes | Yes | Yes | 0.82 | 0.77 | 0.93 |

**Table 6. Full performance results by class**

| | Without Imputation | | | With Imputation | | |
|--|-------------------|--|--|-----------------|--|--|
| | Accuracy | F-Measure (yes) | F-Measure (no) | Accuracy | F-Measure (yes) | F-Measure (no) |
| **kNN-TSC-FW (left) kNN-TSC-FIW (right)** | 0.6 | 0.40 | 0.72 | 0.78 | 0.66 | 0.89 |
| **Lee et al.** | 0.52 | 0.27 | 0.67 | 0.72 | 0.56 | 0.81 |
| **Manual weights** | 0.57 | 0.33 | 0.71 | 0.71 | 0.59 | 0.81 |
| **Chi Square** | 0.50 | 0.24 | 0.64 | 0.68 | 0.52 | 0.77 |
| **Information Gain** | 0.53 | 0.28 | 0.68 | 0.71 | 0.55 | 0.81 |
| **Gini Index** | 0.55 | 0.30 | 0.7 | 0.70 | 0.58 | 0.81 |
| **No Feature Weights** | 0.44 | 0.19 | 0.62 | 0.49 | 0.27 | 0.64 |

**Table 7. Comparison of different feature weighting techniques on the proposed method**

This table shows the advantage of the selected Gradient Descent FW over methods that assign weights based on pre-calculated values as well as domain knowledge based weight assignment (i.e. Manual weights). The performance of a model without the proposed adjacency-based imputation on the left-hand side of Table 7 is significantly lower than the same model with imputation on the right-hand side, providing evidences for the effectiveness of the proposed minimalist imputation method.

Table 8 compares the performance of kNN-TSC-FIWC using different change point detection (CPD) methods including parametric (M-G and V-G) and non-parametric (L-NP, S-NP, and LS-NP) CPD methods as well as the change point detection method proposed by Lin et al. (2014). Although non-parametric CPD approaches perform better than parametric CPD approaches, none of them nor the CPD method proposed by Lin et al. [16] could outperform kNN-TSC-FIWC. The comparison provides evidences that changes in patient time series cannot be detected without considering domain-based patient status categories. In addition, our post process analysis shows that patients with higher number of DCHP are more likely to die due to their unstable situation. Patients with negative last change points which indicate declining health status dominate the early death class. These patterns show the importance of the proposed change point detection features.

| Change Point Detection Method | Accuracy | F-Measure (yes) | F-Measure (no) |
|-------------------------------|----------|-----------------|----------------|
| **kNN-TSC-FIWC** | 0.82 | 0.77 | 0.93 |
| **Lin et al.** | 0.75 | 0.63 | 0.86 |
| **L-NP** | 0.79 | 0.68 | 0.9 |
| **S-NP** | 0.75 | 0.63 | 0.76 |
| **LS-NP** | 0.8 | 0.7 | 0.91 |
| **M-G** | 0.75 | 0.62 | 0.84 |
| **V-G** | 0.75 | 0.62 | 0.84 |

**Table 8. Comparison of kNN-TSC-FIWC using different Change Point Detection Methods**

| Window Size | Num of Patients | Died | Without Change Point | | | With Change Point | | |
|---|---|---|---|---|---|---|---|---|
| | | | Accuracy | F-Measure (yes) | F-Measure (no) | Accuracy | F-Measure (yes) | F-Measure (no) |
| **24** | 33625 | 5102 (15%) | 0.72 | 0.60 | 0.81 | 0.78 | 0.67 | 0.9 |
| **48** | 22561 | 4068 (18%) | 0.78 | 0.66 | 0.89 | 0.82 | 0.77 | 0.93 |
| **72** | 15913 | 3389 (21%) | 0.8 | 0.7 | 0.91 | 0.84 | 0.82 | 0.94 |
| **96** | 11976 | 2903 (24%) | 0.75 | 0.63 | 0.86 | 0.82 | 0.75 | 0.92 |
| **12** | 9485 | 2647 (27%) | 0.74 | 0.61 | 0.83 | 0.81 | 0.72 | 0.92 |

**Table 9. Comparison of kNN-TSC-FIWC using different prediction windows**

## Contributions and Limitations

This study makes several contributions to the patient time series classification and the early ICU mortality prediction research fields:

Based on the patient similarity theory, the study evaluates the effectiveness of the similarity-based patient time series classification approach.

The study further evaluates and identifies effective fine-grain PAA temporal abstraction, similarity functions and proposes necessary enhancements via adjacency-based missing value imputation. The study also evaluates the effectiveness of the gradient decent feature weight assignment approach for reducing temporal dimensions and improving accuracy.

To the best of our knowledge, the study is the first to propose directional patient status change point detection to extract effective features for patient time series classification.

The study contributes to solutions to an important healthcare predictive problem – early ICU mortality prediction by significantly improving prediction accuracy with a new framework that embeds effective extant methods and new enhancements. Both intensive caregivers and patients' families can benefit from this framework with the crucial decision on aggressive or supportive treatment. Also, unexpected deaths, which are still common despite evidence that patients often show signs of clinical deterioration hours in advance, can be detected.

The main limitations of this study include the use of a single data set for evaluation, the difficulty of explaining a *k*NN model, and the need to examine additional methods appropriate for time series classification. Future research should pursue along the directions that could address these limitations.